\documentclass[11pt,a4paper]{article}
\usepackage[hyperref]{acl2019}
\usepackage{times}
\usepackage{latexsym}
\usepackage{graphicx}
\usepackage{url}

\aclfinalcopy 

\title{Do Neural Dialog Systems Use the Conversation History Effectively? \\ An Empirical Study}

\author{\qquad Chinnadhurai Sankar$^{1,2,4}$\thanks{ ~~Corresponding author: chinnadhurai@gmail.com}\qquad~~ Sandeep Subramanian$^{1,2,5}$\\\\ \qquad\qquad \textbf{Christopher Pal}$^{1,3,5}$ ~~~~~~~\qquad\textbf{Sarath Chandar}$^{1,2,4}$ \qquad\qquad \textbf{Yoshua Bengio }$^{1,2}$\\\\
    $^1$\small{Mila} \qquad
    $^2$\small{Universit\'e de Montr\'eal} \qquad 
    $^3$\small{\'Ecole Polytechnique de Montr\'eal} \qquad \\
    $^4$\small{Google Research, Brain Team} \qquad 
    $^5$\small{Element AI, Montr\'eal}
}

\date{}

\begin{document}
\maketitle
\begin{abstract}
  Neural generative models have been become increasingly popular when building conversational agents. They offer flexibility, can be easily adapted to new domains, and require minimal domain engineering. A common criticism of these systems is that they seldom understand or use the available dialog history effectively. In this paper, we take an empirical approach to understanding how these models use the available dialog history by studying the sensitivity of the models to artificially introduced \textit{unnatural} changes or perturbations to their context at test time. We experiment with 10 different types of perturbations on 4 multi-turn dialog datasets and find that commonly used neural dialog architectures like recurrent and transformer-based seq2seq models are rarely sensitive to most perturbations such as missing or reordering utterances, shuffling words, etc. Also, by open-sourcing our code, we believe that it will serve as a useful diagnostic tool for evaluating dialog systems in the future \footnote{Code is available at \url{https://github.com/chinnadhurai/ParlAI/}}.
  
\end{abstract}

\begin{table*}[h!]
\small
\begin{tabular}{|c|p{7.3cm}|p{7.3cm}|}
\hline
 & \textbf{No Perturbations} & \textbf{Token shuffling} \\
\hline
1 & Good afternoon ! Can I help you ? & I afternoon help you Good ? ! Can \\
2 & Could you show me where the Chinesc-style clothing is located ? I want to buy a silk coat & the located Chinesc-style where is show a . buy you ? I clothing want coat silk me Could to \\
3 & This way , please . Here they are . They're all handmade . & are handmade . way please This all Here they . , They're . \\
\hline
4 & \textbf{Model Response}: How much is it ? & \textbf{Model Response}: How much is it ? \\
\hline
\end{tabular}
\caption{An example of an LSTM seq2seq model with attention's insensitivity to shuffling of words in the dialog history on the DailyDialog dataset.}
\label{table:example}
\end{table*}

\section{Introduction}
With recent advancements in generative models of text \cite{wu2016google, vaswani2017attention, radford2018improving}, neural approaches to building chit-chat and goal-oriented conversational agents \cite{sordoniDialog, vinyals2015neural, hred, jasonWestonGoalOriented, serban2017deep} has gained popularity with the hope that advancements in tasks like machine translation \cite{bahdanau2014neural}, abstractive summarization \cite{see2017get} should translate to dialog systems as well. While these models have demonstrated the ability to generate fluent responses, they still lack the ability to ``understand'' and process the dialog history to produce coherent and interesting responses. They often produce boring and repetitive responses like ``Thank you.'' \cite{li_diversity,serban2016hierarchical} or meander away from the topic of conversation. This has been often attributed to the manner and extent to which these models use the dialog history when generating responses. However, there has been little empirical investigation to validate these speculations.

In this work, we take a step in that direction and confirm some of these speculations, showing that models do not make use of a lot of the information available to it, by subjecting the dialog history to a variety of synthetic perturbations. We then empirically observe how recurrent \cite{sutskever2014sequence} and transformer-based \cite{vaswani2017attention} sequence-to-sequence (seq2seq) models respond to these changes. The central premise of this work is that \textit{models make minimal use of certain types of information if they are insensitive to perturbations that destroy them}. Worryingly, we find that 1) both recurrent and transformer-based seq2seq models are insensitive to most kinds of perturbations considered in this work 2) both are particularly insensitive even to extreme perturbations such as randomly shuffling or reversing words within every utterance in the conversation history (see Table \ref{table:example}) and 3) recurrent models are more sensitive to the ordering of utterances within the dialog history, suggesting that they could be modeling conversation dynamics better than transformers. 

\section{Related Work}
Since this work aims at investigating and gaining an understanding of the kinds of information a generative neural response model learns to use, the most relevant pieces of work are where similar analyses have been carried out to understand the behavior of neural models in other settings. An investigation into how LSTM based \textit{unconditional} language models use available context was carried out by \citet{khandelwal2018sharp}. They empirically demonstrate that models are sensitive to perturbations only in the nearby context and typically use only about 150 words of context. On the other hand, in conditional language modeling tasks like machine translation, models are adversely affected by both synthetic and natural noise introduced anywhere in the input \citep{belinkov2017synthetic}. Understanding what information is learned or contained in the representations of neural networks has also been studied by ``probing'' them with linear or deep models \cite{adi2016fine, subramanian2018learning, conneau2018you}.

Several works have recently pointed out the presence of annotation artifacts in common text and multi-modal benchmarks. For example, \citet{gururangan2018annotation} demonstrate that hypothesis-only baselines for natural language inference obtain results \textit{significantly} better than random guessing. \citet{kaushik2018much} report that reading comprehension systems can often ignore the entire question or use only the last sentence of a document to answer questions. \citet{anand2018blindfold} show that an agent that does not navigate or even see the world around it can answer questions about it as well as one that does. These pieces of work suggest that while neural methods have the potential to learn the task specified, its design could lead them to do so in a manner that doesn't use all of the available information within the task.

Recent work has also investigated the inductive biases that different sequence models learn. For example, \citet{tran2018importance} find that recurrent models are better at modeling hierarchical structure while \citet{tang2018self} find that feedforward architectures like the transformer and convolutional models are not better than RNNs at modeling long-distance agreement. Transformers however excel at word-sense disambiguation. We analyze whether the choice of architecture and the use of an attention mechanism affect the way in which dialog systems use information available to them.

\begin{figure*}
\centering
\includegraphics[scale=0.36]{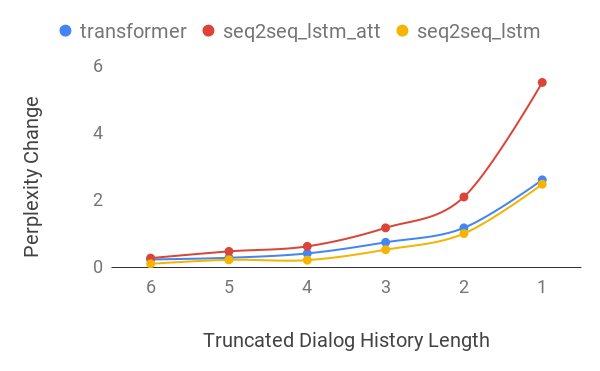}
\includegraphics[scale=0.36]{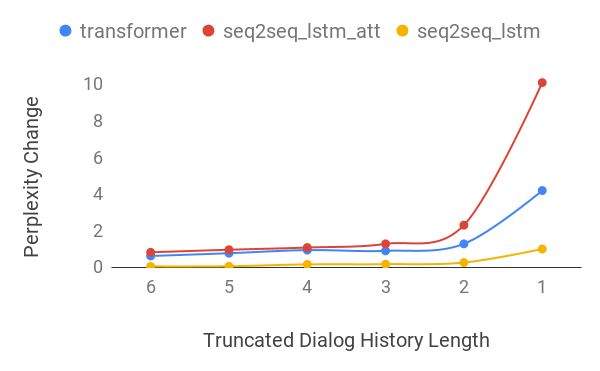}
\caption{The increase in perplexity for different models when only presented with the $k$ most recent utterances from the dialog history for Dailydialog (left) and bAbI dialog (right) datasets. Recurrent models with attention fare better than transformers, since they use more of the conversation history.}
\label{fig:daily_babi5}
\end{figure*}

\section{Experimental Setup}
Following the recent line of work on generative dialog systems, we treat the problem of generating an appropriate response given a conversation history as a conditional language modeling problem. Specifically we want to learn a conditional probability distribution $P_{\theta}(y|x)$ where $y$ is a reasonable response given the conversation history $x$. The conversation history is typically represented as a sequence of utterances $\mathbf{x_1, x_2, \ldots x_n}$, where each utterance $\mathbf{x_i}$ itself is comprised of a sequence of words $x_{i_1}, x_{i_2} \ldots x_{i_k}$. The response $y$ is a single utterance also comprised of a sequence of words $y_1, y_2 \ldots y_m$. The overall conditional probability is factorized autoregressively as $$P_{\theta}(\mathbf{y}|\mathbf{x}) = \prod_{i=1}^{n} P_{\theta}(y_i|y_{<i},\mathbf{x_1 \ldots x_n)}$$

$P_{\theta}$, in this work, is parameterized by a recurrent or transformer-based seq2seq model. The \textit{crux of this work is to study how the learned probability distribution behaves as we artificially perturb the conversation history} $\mathbf{x_1, \ldots x_n}$. We measure behavior by looking at how much the per-token perplexity increases under these changes. For example, one could think of shuffling the order in which $\mathbf{x_1 \ldots x_n}$ is presented to the model and observe how much the perplexity of $\mathbf{y}$ under the model increases. If the increase is only minimal, we can conclude that the ordering of $\mathbf{x_1 \ldots x_n}$ isn't informative to the model. For a complete list of perturbations considered in this work, please refer to Section \ref{sec:perturbations}. All models are trained without any perturbations and sensitivity is studied \textit{only at test time}.

\subsection{Datasets}
We experiment with four multi-turn dialog datasets.

\paragraph{bAbI dialog} is a synthetic goal-oriented multi-turn dataset \citep{jasonWestonGoalOriented} consisting of 5 different tasks for restaurant booking with increasing levels of complexity.  We consider Task 5 in our experiments since it is the hardest and is a union of all four tasks.  It contains $1k$ dialogs with an average of $13$ user utterances per dialog.

\paragraph{Persona Chat} is an open domain dataset \citep{zhang2018personalizing} with multi-turn chit-chat conversations between turkers who are each assigned a ``persona'' at random. It comprises of $10.9 k$ dialogs with an average of $14.8$ turns per dialog.

\paragraph{Dailydialog} is an open domain dataset \citep{li2017dailydialog} which consists of dialogs that resemble day-to-day conversations across multiple topics. It comprises of $13k$ dialogs with an average of $7.9$ turns per dialog.

\paragraph{MutualFriends} is a multi-turn goal-oriented dataset \citep{mutualFriends} where two agents must discover which friend of theirs is mutual based on the friends' attributes. It contains $11k$ dialogs with an average of $11.41$ utterances per dialog.

\begin{table*}[ht]
\scriptsize
\begin{tabular}{l|c|p{0.8cm}p{0.8cm}p{0.8cm}p{0.8cm}p{0.8cm}|p{0.8cm}p{0.8cm}p{0.8cm}p{0.8cm}p{0.8cm}}
\hline
Models & Test PPL & Only Last & Shuf & Rev & Drop First & Drop Last & Word Drop & Verb Drop & Noun Drop & Word Shuf & Word Rev \\
\hline
\multicolumn{2}{c|}{} & \multicolumn{5}{c|}{Utterance level perturbations $\;$ ($\;\Delta$ $PPL_{[\sigma]}\;$)} & \multicolumn{5}{c}{Word level perturbations  $\;$ ($\;\Delta$ $PPL_{[\sigma]}\;$)}  \\
\hline
\multicolumn{12}{c}{\textbf{DailyDialog}}\\
\hline
seq2seq\_lstm & 32.90$_{[1.40]}$ & 1.70$_{[0.41]}$ & \textbf{3.35$_{[0.38]}$} & \textbf{4.04$_{[0.28]}$} & 0.13$_{[0.04]}$ & \textbf{5.08$_{[0.79]}$} & 1.58$_{[0.15]}$ & 0.87$_{[0.08]}$ & 1.06$_{[0.28]}$ & \textbf{3.37$_{[0.33]}$} & 3.10$_{[0.45]}$ \\
seq2seq\_lstm\_att & 29.65$_{[1.10]}$ & \textbf{4.76$_{[0.39]}$} & 2.54$_{[0.24]}$ & 3.31$_{[0.49]}$ & \textbf{0.32$_{[0.03]}$} & 4.84$_{[0.42]}$ & \textbf{2.03$_{[0.25]}$} & \textbf{1.37$_{[0.29]}$} & 2.22$_{[0.22]}$ & 2.82$_{[0.31]}$ & \textbf{3.29$_{[0.25]}$} \\
transformer & \textbf{28.73$_{[1.30]}$} & 3.28$_{[1.37]}$ & 0.82$_{[0.40]}$ & 1.25$_{[0.62]}$ & 0.27$_{[0.19]}$ & 2.43$_{[0.83]}$ & 1.20$_{[0.69]}$ & 0.63$_{[0.17]}$ & \textbf{2.60$_{[0.98]}$} & 0.15$_{[0.08]}$ & 0.26$_{[0.18]}$ \\
\hline
\multicolumn{12}{c}{\textbf{Persona Chat}}\\
\hline  
seq2seq\_lstm & 43.24$_{[0.99]}$ & 3.27$_{[0.13]}$ & 6.29$_{[0.48]}$ & \textbf{13.11$_{[1.22]}$} & 0.47$_{[0.21]}$ & \textbf{6.10$_{[0.46]}$} & 1.81$_{[0.25]}$ & 0.68$_{[0.19]}$ & 0.75$_{[0.15]}$ & 1.29$_{[0.17]}$ & 1.95$_{[0.20]}$ \\
seq2seq\_lstm\_att & 42.90$_{[1.76]}$ & \textbf{4.44$_{[0.81]}$} & \textbf{6.70$_{[0.67]}$} & 11.61$_{[0.75]}$ & \textbf{2.99$_{[2.24]}$} & 5.58$_{[0.45]}$ & \textbf{2.47$_{[0.67]}$} & \textbf{1.11$_{[0.27]}$} & \textbf{1.20$_{[0.23]}$} & \textbf{2.03$_{[0.46]}$} & \textbf{2.39$_{[0.31]}$} \\
transformer & \textbf{40.78$_{[0.31]}$} & 1.90$_{[0.08]}$ & 1.22$_{[0.22]}$ & 1.41$_{[0.54]}$ & $-$0.1$_{[0.07]}$ & 1.59$_{[0.39]}$ & 0.54$_{[0.08]}$ & 0.40$_{[0.00]}$ & 0.32$_{[0.18]}$ & 0.01$_{[0.01]}$ & 0.00$_{[0.06]}$ \\
\hline
\multicolumn{12}{c}{\textbf{MutualFriends}}\\
\hline  
seq2seq\_lstm & 14.17$_{[0.29]}$ & 1.44$_{[0.86]}$ & \textbf{1.42$_{[0.25]}$} & 1.24$_{[0.34]}$ & 0.00$_{[0.00]}$ & 0.76$_{[0.10]}$ & 0.28$_{[0.11]}$ & 0.00$_{[0.03]}$ & 0.61$_{[0.39]}$ & 0.31$_{[0.25]}$ & 0.56$_{[0.39]}$ \\
seq2seq\_lstm\_att & \textbf{10.60$_{[0.21]}$} & \textbf{32.13$_{[4.08]}$} & 1.24$_{[0.19]}$ & 1.06$_{[0.24]}$ & 0.08$_{[0.03]}$ & \textbf{1.35$_{[0.15]}$} & \textbf{1.56$_{[0.20]}$} & 0.15$_{[0.07]}$ & \textbf{3.28$_{[0.38]}$} & \textbf{2.35$_{[0.22]}$} & \textbf{4.59$_{[0.46]}$} \\
transformer & 10.63$_{[0.03]}$ & 20.11$_{[0.67]}$ & 1.06$_{[0.16]}$ & \textbf{1.62$_{[0.44]}$} & \textbf{0.12$_{[0.03]}$} & 0.81$_{[0.09]}$ & 0.75$_{[0.05]}$ & \textbf{0.16$_{[0.02]}$} & 1.50$_{[0.12]}$ & 0.07$_{[0.01]}$ & 0.13$_{[0.04]}$ \\
\hline
\multicolumn{12}{c}{\textbf{bAbi dailog: Task5}}\\
\hline  
seq2seq\_lstm & 1.28$_{[0.02]}$ & 1.31$_{[0.50]}$ & \textbf{43.61$_{[15.9]}$} & \textbf{40.99$_{[9.38]}$} & 0.00$_{[0.00]}$ & 4.28$_{[1.90]}$ & 0.38$_{[0.11]}$ & 0.01$_{[0.00]}$ & 0.10$_{[0.06]}$ & 0.09$_{[0.02]}$ & 0.42$_{[0.38]}$ \\
seq2seq\_lstm\_att & \textbf{1.06$_{[0.02]}$} & \textbf{9.14$_{[1.28]}$} & 41.21$_{[8.03]}$ & 34.32$_{[10.7]}$ & 0.00$_{[0.00]}$ & \textbf{6.75$_{[1.86]}$} & \textbf{0.64$_{[0.07]}$} & 0.03$_{[0.03]}$ & 0.22$_{[0.04]}$ & \textbf{0.25$_{[0.01]}$} & \textbf{1.10$_{[0.80]}$} \\
transformer & 1.07$_{[0.00]}$ & 4.06$_{[0.33]}$ & 0.38$_{[0.02]}$ & 0.62$_{[0.02]}$ & 0.00$_{[0.00]}$ & 0.21$_{[0.02]}$ & 0.36$_{[0.02]}$ & \textbf{0.25$_{[0.06]}$} & \textbf{0.37$_{[0.06]}$} & 0.00$_{[0.00]}$ & 0.00$_{[0.00]}$ \\
\hline
\end{tabular}
\caption{Model performance across multiple datasets and sensitivity to different perturbations. Columns 1 \& 2 report the test set perplexity (without perturbations) of different models. Columns 3-12 report the \textbf{increase} in perplexity when models are subjected to different perturbations. The mean ($\mu$) and standard deviation $[\sigma]$ across 5 runs are reported. The \textit{Only Last} column presents models with \textbf{only} the last utterance from the dialog history. The model that exhibits the highest sensitivity (higher the better) to a particular perturbation on a dataset is in bold. \textit{seq2seq\_lstm\_att} are the most sensitive models \textbf{24/40} times, while transformers are the least with \textbf{6/40} times.}
\label{table:delta_ppl}
\end{table*}

\subsection{Types of Perturbations}
\label{sec:perturbations}
We experimented with several types of perturbation operations at the utterance and word (token) levels. All perturbations are applied in isolation.

\paragraph{Utterance-level perturbations} We consider the following operations 1) \textit{Shuf} that shuffles the sequence of utterances in the dialog history, 2) \textit{Rev} that reverses the order of utterances in the history (but maintains word order within each utterance) 3) \textit{Drop} that completely drops certain utterances and 4) \textit{Truncate} that truncates the dialog history to contain only the $k$ most recent utterances where $k \leq n$, where n is the length of dialog history.

\paragraph{Word-level perturbations} We consider similar operations but at the word level within \textbf{every} utterance 1) \textit{word-shuffle} that randomly shuffles the words within an utterance 2) \textit{reverse} that reverses the ordering of words, 3) \textit{word-drop} that drops 30\% of the words uniformly 4) \textit{noun-drop} that drops \textit{all} nouns, 5) \textit{verb-drop} that drops \textit{all} verbs.

\subsection{Models}
\label{section:models}
We experimented with two different classes of models - recurrent and transformer-based sequence-to-sequence generative models. All data loading, model implementations and evaluations were done using the ParlAI framework.
We used the default hyper-parameters for all the models as specified in ParlAI.

\paragraph{Recurrent Models} We trained a seq2seq (\textit{seq2seq\_lstm}) model where the encoder and decoder are parameterized as LSTMs \cite{lstm1997}. We also experiment with using decoders that use an attention mechanism (\textit{seq2seq\_lstm\_att}) \cite{bahdanau2014neural}. The encoder and decoder LSTMs have 2 layers with 128 dimensional hidden states with a dropout rate of 0.1.

\paragraph{Transformer} Our transformer \cite{vaswani2017attention} model uses 300 dimensional embeddings and hidden states, 2 layers and 2 attention heads with no dropout. This model is significantly smaller than the ones typically used in machine translation since we found that the model that resembled \citet{vaswani2017attention} significantly overfit on all our datasets.

While the models considered in this work might not be state-of-the-art on the datasets considered, we believe these models are still competitive and used commonly enough at least as baselines, that the community will benefit by understanding their behavior. In this paper, we use early stopping with a patience of $10$ on the validation set to save our best model. All models achieve close to the perplexity numbers reported for generative seq2seq models in their respective papers. 

\section{Results \& Discussion}
Our results are presented in Table \ref{table:delta_ppl} and Figure \ref{fig:daily_babi5}. Table \ref{table:delta_ppl} reports the perplexities of different models on test set in the second column, followed by the \textbf{increase} in perplexity when the dialog history is perturbed using the method specified in the column header. Rows correspond to models trained on different datasets. Figure \ref{fig:daily_babi5} presents the change in perplexity for models when presented only with the $k$ most recent utterances from the dialog history. 

We make the following observations:
\begin{enumerate}
    \item Models tend to show only tiny changes in perplexity in most cases, even under extreme changes to the dialog history, suggesting that they use far from all the information that is available to them.
    \item Transformers are insensitive to word-reordering, indicating that they could be learning bag-of-words like representations.
    \item The use of an attention mechanism in \textit{seq2seq\_lstm\_att} and transformers makes these models use more information from earlier parts of the conversation than vanilla seq2seq models as seen from increases in perplexity when using only the last utterance.
    \item While transformers converge faster and to lower test perplexities, they don't seem to capture the conversational dynamics across utterances in the dialog history and are less sensitive to perturbations that scramble this structure than recurrent models.
\end{enumerate}


%

%

\label{sec:res_dicussion}
\section{Conclusion}
This work studies the behaviour of generative neural dialog systems in the presence of synthetically introduced perturbations to the dialog history, that it conditions on. We find that both recurrent and transformer-based seq2seq models are not significantly affected even by drastic and unnatural modifications to the dialog history. We also find subtle differences between the way in which recurrent and transformer-based models use available context. By open-sourcing our code, we believe this paradigm of studying model behavior by introducing perturbations that destroys different kinds of structure present within the dialog history can be a useful diagnostic tool. We also foresee this paradigm being useful when building new dialog datasets to understand the kinds of information models use to solve them.

\section*{Acknowledgements}
We would like to acknowledge NVIDIA for donating GPUs and a DGX-1 computer used in this work. We would also like to thank the anonymous reviewers for their constructive feedback. Our code is available at \url{https://github.com/chinnadhurai/ParlAI/}. 

\bibliography{acl2019}
\bibliographystyle{acl_natbib}

\end{document}